\documentclass[letterpaper, 10pt, conference]{ieeeconf}

\IEEEoverridecommandlockouts

\usepackage{soul}
\usepackage{tabularx}
\usepackage{graphicx} 
\usepackage{times} 
\usepackage{amsmath} 
\usepackage{amssymb} 
\usepackage{algorithm}
\usepackage{float}
\usepackage[noend]{algpseudocode}
\usepackage{multirow,color}
\usepackage{cite}
\usepackage{algpseudocode}
\usepackage{varwidth}
\usepackage{subfig}
\usepackage{booktabs}
\usepackage{gensymb}
\usepackage[hyphens]{url}
\usepackage[export]{adjustbox}
\usepackage[font=footnotesize]{caption}
\let\labelindent\relax
\usepackage[inline]{enumitem} 
\usepackage{graphicx,tipa}
\usepackage{stfloats}

\usepackage{color} 
\usepackage{siunitx}
\usepackage[hidelinks]{hyperref}
\usepackage{balance}

\makeatletter
\def\BState{\State\hskip-\ALG@thistlm}
\makeatother

\title{\LARGE \bf {Design and Development of a Robotic Transcatheter Delivery System for Aortic Valve Replacement}}
\author{Harith S. Gallage, Bailey F. De Sousa, Benjamin I. Chesnik, Chaikel G. Brownstein, Anson Paul, \\and Ronghuai Qi*, \textit{Member IEEE}
\thanks{This work was supported in part by the Start-Up Fund of the University of Nevada, Las Vegas.}
\thanks{The authors are with the Robotics and Healthcare Systems (RoboHS) Laboratory, Department of Mechanical Engineering, University of Nevada, Las Vegas (UNLV), Las Vegas, NV, USA.}
\thanks{*Ronghuai Qi is the corresponding author
(ronghuai.qi@unlv.edu).}}

\begin{document}
\maketitle
\thispagestyle{empty}
\pagestyle{empty}
\begin{abstract}
Minimally invasive transcatheter approaches are increasingly adopted for aortic stenosis treatment, where optimal commissural and coronary alignment is important. Achieving precise alignment remains clinically challenging, even with contemporary robotic transcatheter aortic valve replacement (TAVR) devices, as this task is still performed manually. This paper proposes the development of a robotic transcatheter delivery system featuring an omnidirectional bending joint and an actuation system designed to enhance positional accuracy and precision in TAVR procedures. The preliminary experimental results validate the functionality of this novel robotic system. 

\end{abstract}
\section{Introduction}\label{sec:intro}
Aortic stenosis is a serious and common condition among the elderly in the U.S. TAVR has become a leading minimally invasive treatment of aortic valve disease which delivers prosthetic valve typically via transfemoral access \cite{7aortic}. As TAVR is used, achieving high-precision valve deployment is critical to ensuring optimal hemodynamics and preventing complications, including coronary obstruction \cite{9predictive}. This proposed design represents an advancement over \cite{toward}, increasing two to four tendons to enable omnidirectional bending, thereby enhancing the flexibility and maneuverability of the robotic catheter within vascular structures. The bending joint assembly comprises a stack of eight concentrically aligned rings interconnected by four tendons which are superelastic Nitinol wires (diameter: 0.16\,mm). The bending joint is actuated with incorporating a commercially available Edwards balloon catheter (Edwards Lifesciences, USA), which has a diameter of 3.2\,mm and length of 130\,cm.

\section{Methods and Results}\label{sec:result}
\begin{figure}
    \centering
    \vspace{-0.2cm}
    \includegraphics[width=0.75\linewidth]{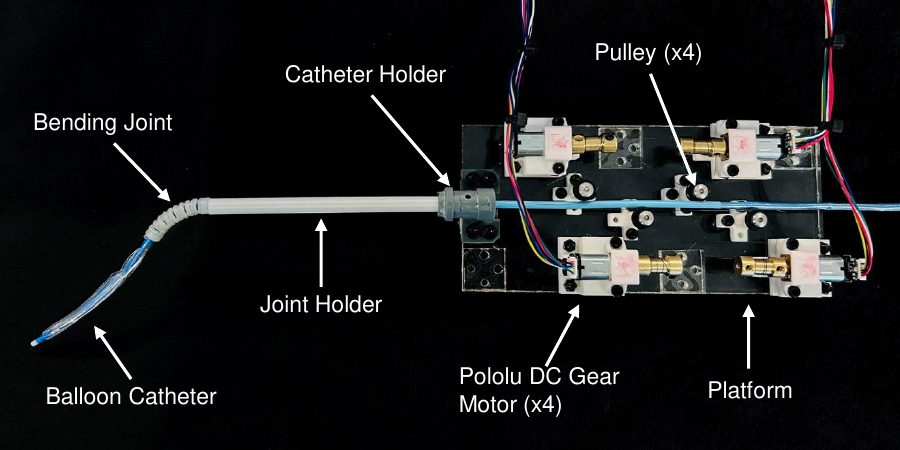}
    \caption{Omnidirectional bending joint with actuation system.}
    \label{fig:Actuation System}
\end{figure}

The bending joint rings were fabricated using Nylon 6/6 material and has outer diameter of 7\,mm. The proposed design features four tendons arranged at equal radial spacing (90° apart) along a constant radius of 2.5\,mm. This configuration enables omnidirectional bending through a single constant-curvature section. To maintain constant curvature, the system coordinates paired tendon actuation, where opposing tendons simultaneously pull and release equal lengths to produce controlled bending which are actuated by Pololu DC gear motors. Experiment is carried out by aforementioned joint control method. The actuation system has a compact design measuring 183 × 80 × 38\,mm as shown in Fig. \ref{fig:Actuation System}.

\begin{figure}
    \centering
    \vspace{-0.2cm}
    \includegraphics[width=0.75\linewidth]{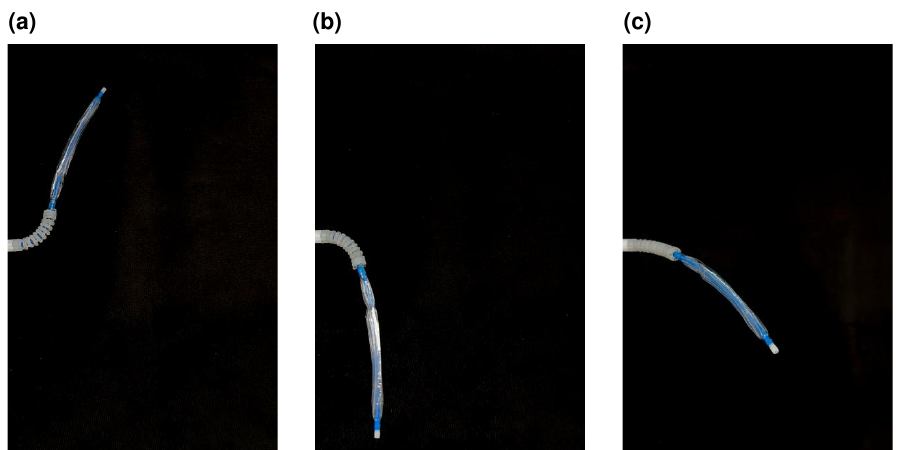}
    \caption{Robot motion validation experiments. (a) $\sim\ang{90}$ upward bend, (b) $\sim\ang{90}$ downward bend, and (c) omnidirectional bend.}
    \label{fig:Bending}
    \vspace{-0.5cm}
\end{figure}

The developed bending joint achieves maximum bending angles of $\sim\ang{90}$ in horizontal plane in both directions, as demonstrated in Fig. \ref{fig:Bending}(a) and Fig. \ref{fig:Bending}(b). Furthermore, the joint successfully exhibits omnidirectional bending capability, as shown in Fig. \ref{fig:Bending}(c). These results demonstrate the potential to enhance existing robotic transcatheter systems through integration of this novel bending joint.

\section{Conclusions and Future Work}\label{sec:Con}
In this paper, we proposed and fabricated a novel omnidirectional bending joint with an actuation system capable of actuating the commercial balloon catheter for TAVR. The maximum bending angle of approximately $\ang{90}$ and omnidirectional bending are successfully achieved in experimental results and validate the design's functionality and its potential for future development. Our future work will be focused on hysteresis modeling, kinematic modeling, and experimental validation using patient-specific phantoms.

\bibliographystyle{IEEEtran}
\bibliography{references}

\end{document}